\documentclass{article} 
\pdfoutput=1 
\usepackage[utf8]{inputenc}
\usepackage[T1]{fontenc}
\usepackage{iclr2017_conference}
\usepackage[hidelinks]{hyperref}
\usepackage{graphicx}
\usepackage{subfigure}
\usepackage{hyperref}
\usepackage{url}
\usepackage{times}
\usepackage{epsfig}
\usepackage{graphicx}
\usepackage{amsmath}
\usepackage{amssymb}
\usepackage{multirow}
\usepackage{caption}
\usepackage{minipage} 
\usepackage{url}
\usepackage{booktabs}

\title{Recurrent Batch Normalization}

\author{
Tim Cooijmans, Nicolas Ballas, C\'esar Laurent, Çağlar Gülçehre \& Aaron Courville \\
MILA - Universit\'e de Montr\'eal \\
\texttt{firstname.lastname@umontreal.ca} \\
}

\usepackage{amsfonts}
\usepackage{amsmath}

\newcommand{\vect}[1]{\mathbf{#1}}
\newcommand{\mat}[1]{\mathbf{#1}}

\newcommand{\ewprod}{\odot}
\newcommand{\reals}{\mathbb{R}}

\iclrfinalcopy

\begin{document}

\maketitle

\begin{abstract}
We propose a reparameterization of LSTM that brings the benefits of batch normalization to recurrent neural networks.
Whereas previous works only apply batch normalization to the input-to-hidden transformation of RNNs,
we demonstrate that it is both possible and beneficial to batch-normalize the hidden-to-hidden transition,
thereby reducing internal covariate shift between time steps.

We evaluate our proposal on various sequential problems such as sequence classification, language modeling and question answering.
Our empirical results show that our batch-normalized LSTM consistently leads to faster convergence and improved generalization.
\end{abstract}

\section{Introduction}

Recurrent neural network architectures such as LSTM~\citep{lstm} and GRU~\citep{cho2014learning} have recently exhibited
state-of-the-art performance on a wide range of complex sequential problems including speech recognition~\cite{baidu},
machine translation~\citep{bahdanau2014neural} and image and video captioning~\citep{xu2015show,yao2015describing}.
Top-performing models, however, are based on very high-capacity networks that are computationally intensive and costly to train.
Effective optimization of recurrent neural networks is thus an active area of study~\citep{pascanudifficulty,hessianfree,ollivier}.

It is well-known that for deep feed-forward neural networks, covariate
shift~\citep{shimodaira2000improving,batchnorm} degrades the efficiency of training.  Covariate
shift is a change in the distribution of the inputs to a model.  This occurs continuously during
training of feed-forward neural networks, where changing the parameters of a layer affects the
distribution of the inputs to all layers above it.  As a result, the upper layers are continually
adapting to the shifting input distribution and unable to learn effectively.  This \emph{internal}
covariate shift~\citep{batchnorm} may play an especially important role in recurrent neural
networks, which resemble very deep feed-forward networks.

Batch normalization~\citep{batchnorm} is a recently proposed technique for controlling the
distributions of feed-forward neural network activations, thereby reducing internal covariate
shift.  It involves standardizing the activations going into each layer, enforcing their means and
variances to be invariant to changes in the parameters of the underlying layers.
This effectively decouples each layer's parameters from those of other layers, leading to a
better-conditioned optimization problem.  Indeed, deep neural networks trained with batch
normalization converge significantly faster and generalize better.

Although batch normalization has demonstrated significant training speed-ups and generalization
benefits in feed-forward networks, it is proven to be difficult to apply in recurrent
architectures~\citep{cesar,baidu}.  It has found limited use in stacked RNNs, where the
normalization is applied ``vertically'', i.e. to the input of each RNN, but not ``horizontally''
between timesteps.  RNNs are deeper in the time direction, and as such batch normalization would
be most beneficial when applied horizontally.  However,~\citet{cesar} hypothesized that applying
batch normalization in this way hurts training because of exploding gradients due to repeated
rescaling.

Our findings run counter to this hypothesis.  We show that it is both possible and highly
beneficial to apply batch normalization in the hidden-to-hidden transition of recurrent models.
In particular, we describe a reparameterization of LSTM
(Section~\ref{sec:recurrent-batch-normalization}) that involves batch normalization and
demonstrate that it is easier to optimize and generalizes better.  In addition, we empirically
analyze the gradient backpropagation and show that proper initialization of the batch
normalization parameters is crucial to avoiding vanishing gradient
(Section~\ref{sec:activation-variance}).  We evaluate our proposal on several sequential problems
and show (Section~\ref{sec:experiments}) that our LSTM reparameterization consistently outperforms
the LSTM baseline across tasks, in terms of both time to convergence and performance.

\citet{liao2016bridging} simultaneously investigated batch normalization in recurrent neural networks,
albeit only for very short sequences (10 steps).
\citet{ba2016layer} independently developed a variant of batch normalization that is
also applicable to recurrent neural networks and delivers similar improvements as 
our method.

\section{Prerequisites}
\label{sec:prerequisites}

\subsection{LSTM}

Long Short-Term Memory (LSTM) networks are an instance of a more general class of recurrent neural
networks (RNNs), which we review briefly in this paper.  Given an input sequence $\mat{X} = ( \vect{x}_1,
\vect{x}_2, \ldots, \vect{x}_T )$, an RNN defines a sequence of hidden states $\vect{h}_t$
according to
\begin{eqnarray}
  \vect{h}_t = \phi(\mat{W}_h \vect{h}_{t-1} + \mat{W}_x  \vect{x}_t + \vect{b}),
\end{eqnarray}
where $\mat{W}_h \in \reals^{d_h \times d_h}, \mat{W}_x \in \reals^{d_x \times d_h}, \vect{b} \in \reals^{d_h}$
and the initial state $\vect{h}_0 \in \reals^{d_h}$ 
are model parameters.
A popular choice for the activation function $\phi(\ \cdot\ )$ is $\tanh$.

RNNs are popular in sequence modeling thanks to their natural ability to process variable-length sequences.
However, training RNNs using first-order stochastic gradient descent (SGD) is notoriously difficult
due to the well-known problem of exploding/vanishing gradients~\citep{bengio1994learning,hochreiter1991untersuchungen,pascanudifficulty}.
Gradient vanishing occurs when states $\vect{h}_t$ are not influenced by small changes in much earlier states $\vect{h}_{\tau}$, $t \ll \tau$,
preventing learning of long-term dependencies in the input data.
Although learning long-term dependencies is fundamentally difficult~\citep{bengio1994learning},
its effects can be mitigated through architectural variations such as LSTM~\citep{lstm}, GRU~\citep{cho2014learning} and $i$RNN/$u$RNN~\citep{le2015simple,urnn}.

In what follows, we focus on the LSTM architecture~\citep{lstm} with recurrent transition given by
\begin{eqnarray}
\left(\begin{array}{ccc}
\tilde{\vect{f}}_t \\
\tilde{\vect{i}}_t \\
\tilde{\vect{o}}_t \\
\tilde{\vect{g}}_t
\end{array}\right)
 &=&
 \mat{W}_h \vect{h}_{t-1} +
 \mat{W}_x \vect{x}_t +
 \vect{b}
 \\
\vect{c}_t &= &\sigma(\tilde{\vect{f}}_t) \ewprod \vect{c}_{t-1} +
\sigma(\tilde{\vect{i}}_t) \ewprod \tanh(\tilde{\vect{g}_t}) \\
\vect{h}_t &= &\sigma(\tilde{\vect{o}}_t) \ewprod \tanh(\vect{c}_t),
\end{eqnarray}
where $\vect{W}_h \in \reals^{d_h \times 4 d_h}, \vect{W}_x \reals^{d_x \times 4 d_h}, \vect{b} \in \reals^{4 d_h}$
and the initial states $\vect{h}_0 \in \reals^{d_h}, \vect{c}_0 \in \reals^{d_h}$ 
are model parameters.
$\sigma$ is the logistic sigmoid function, and the $\ewprod$ operator denotes the Hadamard product.

The LSTM differs from simple RNNs in that it has an additional memory \emph{cell} 
$\vect{c}_t$ whose update is nearly linear which allows the gradient to flow back 
through time more easily. In addition, unlike the RNN which overwrites its content 
at each timestep, the update of the LSTM cell is regulated by a set of gates.
The forget gate $\vect{f}_t$ determines the extent to which information is carried over from the previous timestep,
and the input gate $\vect{i}_t$ controls the flow of information from the current input $\vect{x}_t$.
The output gate $\vect{o}_t$ allows the model to read from the cell. This carefully 
controlled interaction with the cell is what allows the LSTM to robustly retain 
information for long periods of time.

\subsection{Batch Normalization}

\emph{Covariate shift}~\citep{shimodaira2000improving} is a phenomenon in machine learning where
the features presented to a model change in distribution.
In order for learning to succeed in the presence of covariate shift,
the model's parameters must be adjusted not just to learn the concept at hand
but also to adapt to the changing distribution of the inputs.
In deep neural networks, this problem manifests as \emph{internal covariate shift}~\citep{batchnorm},
where changing the parameters of a layer affects the distribution of the inputs to all layers above it.

Batch Normalization~\citep{batchnorm} is a recently proposed network
reparameterization which aims to reduce internal covariate shift.  It does so by
standardizing the activations using empirical estimates of their means and
standard deviations.  However, it does not decorrelate the activations due to
the computationally costly matrix inversion.  The batch normalizing transform
is as follows:

\begin{align}
\mathrm{BN}(\vect{h}; \gamma, \beta) =
  \beta + \gamma \ewprod
  \frac{\vect{h} -   \widehat{\mathbb{E  }}[\vect{h}]}
       {       \sqrt{\widehat{\mathrm{Var}}[\vect{h}] + \epsilon}}
\end{align}

where $\vect{h} \in \reals^d$ is the vector of (pre)activations to be
normalized, $\gamma \in \reals^d, \beta \in \reals^d$ are model parameters that
determine the mean and standard deviation of the normalized activation, and
$\epsilon \in \reals$ is a regularization hyperparameter. The division should
be understood to proceed elementwise.

At training time, the statistics $\mathbb{E}[\vect{h}]$ and
$\mathrm{Var}[\vect{h}]$ are estimated by the sample mean and sample variance
of the current minibatch.  This allows for backpropagation through the
statistics, preserving the convergence properties of stochastic gradient
descent.  During inference, the statistics are typically estimated based on the
entire training set, so as to produce a deterministic prediction.

\section{Batch-Normalized LSTM}
\label{sec:recurrent-batch-normalization}

This section introduces a reparameterization of LSTM that takes advantage of
batch normalization. Contrary to~\citet{cesar, baidu}, we leverage batch
normalization in both the input-to-hidden \emph{and} the hidden-to-hidden
transformations.  We introduce the batch-normalizing transform $\mathrm{BN}(\
\cdot\ ; \gamma, \beta)$ into the LSTM as follows:

\begin{eqnarray}
\left(\begin{array}{ccc}
\tilde{\vect{f}}_t \\
\tilde{\vect{i}}_t \\
\tilde{\vect{o}}_t \\
\tilde{\vect{g}}_t
\end{array}\right)
 &=&
 \mathrm{BN} (\mat{W}_h \vect{h}_{t-1}; \gamma_h, \beta_h) +
 \mathrm{BN} (\mat{W}_x \vect{x}_t   ; \gamma_x, \beta_x) +
 \vect{b}
\\
\vect{c}_t &=& \sigma(\tilde{\vect{f}}_t) \ewprod \vect{c}_{t-1} +
               \sigma(\tilde{\vect{i}}_t) \ewprod \tanh(\tilde{\vect{g}_t}) \\
\vect{h}_t &=& \sigma(\tilde{\vect{o}}_t) \ewprod \tanh(
 \mathrm{BN} (\vect{c}_t; \gamma_c, \beta_c)
)
\end{eqnarray}

In our formulation, we normalize the recurrent term $\mat{W}_h \vect{h}_{t-1}$
and the input term $\mat{W}_x \vect{x}_t$ separately.  Normalizing these terms
individually gives the model better control over the relative contribution of
the terms using the $\gamma_h$ and $\gamma_x$ parameters.  We set $\beta_h =
\beta_x = \vect{0}$ to avoid unnecessary redundancy, instead relying on the
pre-existing parameter vector $\vect{b}$ to account for both biases.  In order
to leave the LSTM dynamics intact and preserve the gradient flow through
$\vect{c}_t$, we do not apply batch normalization in the cell update.

The batch normalization transform relies on batch statistics to standardize the
LSTM activations.  It would seem natural to share the statistics that are used
for normalization across time, just as recurrent neural networks share their
parameters over time.  However, we find that simply averaging statistics over
time severely degrades performance.  Although LSTM activations do converge to a
stationary distribution, we observe that their statistics during the initial
transient differ significantly (see Figure~\ref{fig:popstat_stationarity} in Appendix~\ref{sec:popstat_stationarity}).
Consequently, we recommend using separate statistics for each timestep to
preserve information of the initial transient phase in the
activations.\footnote{ Note that we separate \emph{only} the statistics over
time and not the $\gamma$ and $\beta$ parameters.}

Generalizing the model to sequences longer than those seen during training is
straightforward thanks to the rapid convergence of the activations to their
steady-state distributions (cf. Figure~\ref{fig:popstat_stationarity}).  For
our experiments we estimate the population statistics separately for each
timestep $1, \ldots, T_{max}$ where $T_{max}$ is the length of the longest
training sequence.  When at test time we need to generalize beyond $T_{max}$,
we use the population statistic of time $T_{max}$ for all time steps beyond it.

During training we estimate the statistics across the minibatch, independently
for each timestep.  At test time we use estimates obtained by averaging the
minibatch estimates over the training set.

\section{Initializing $\gamma$ for Gradient Flow}
\label{sec:activation-variance}

Although batch normalization allows for easy control of the pre-activation
variance through the $\gamma$ parameters, common practice is to normalize to
unit variance.  We suspect that the previous difficulties with recurrent batch
normalization reported in~\citet{cesar,baidu} are largely due to improper
initialization of the batch normalization parameters, and $\gamma$ in
particular.  In this section we demonstrate the impact of $\gamma$ on gradient
flow.

\begin{figure}[!ht]
  \center%
  \subfigure[
We visualize the gradient flow through a batch-normalized $\tanh$ RNN as a
function of $\gamma$.  High variance causes vanishing gradient.
]{%
    \label{fig:rnn_grad_prop}
    \includegraphics[width=.45\textwidth]{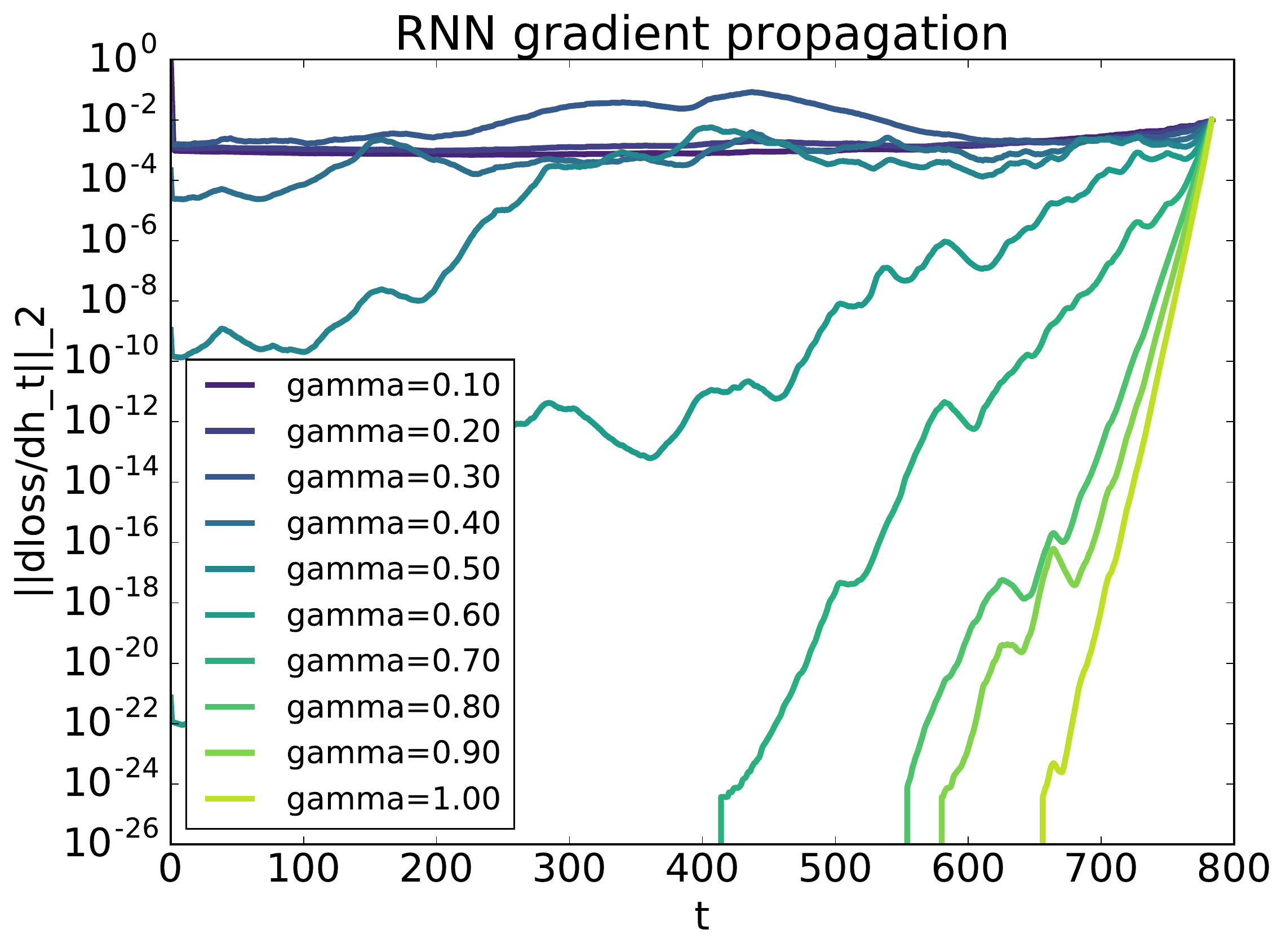}
  }%
  \hspace{2mm}%
  \subfigure[
We show the empirical expected derivative and interquartile range of $\tanh$
nonlinearity as a function of input variance.  High variance causes saturation,
which decreases the expected derivative.
]{%
    \label{fig:tanh_grad}
    \includegraphics[width=.45\textwidth]{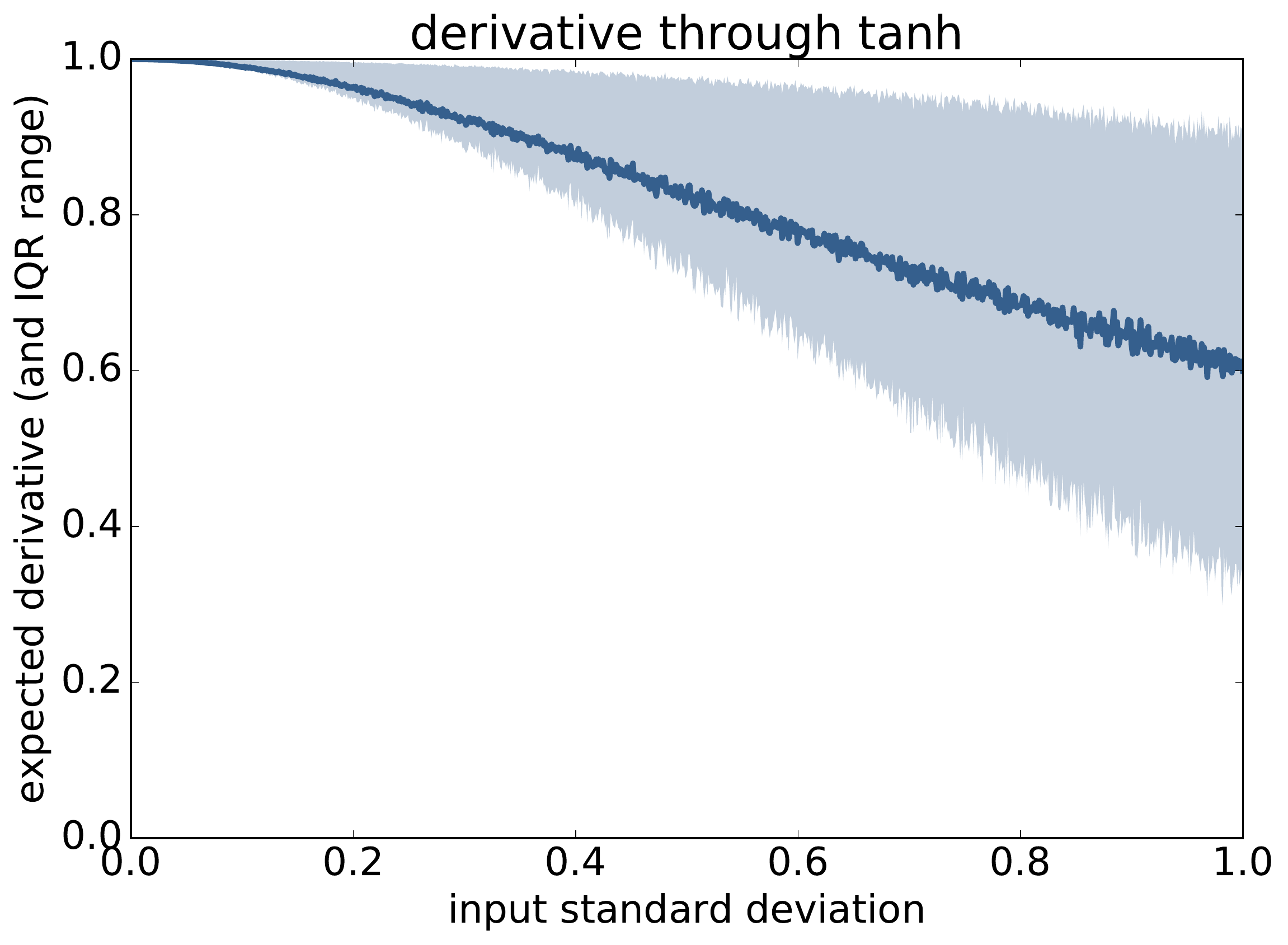}
  }
  \caption{
Influence of pre-activation variance on gradient propagation.
}
  \label{fig:variance}
\end{figure}

In Figure~\ref{fig:rnn_grad_prop}, we show how the pre-activation variance
impacts gradient propagation in a simple RNN on the sequential MNIST task
described in Section~\ref{sec:seqmnist}.  Since backpropagation operates in
reverse, the plot is best read from right to left.  The quantity plotted is the
norm of the gradient of the loss with respect to the hidden state at different
time steps.  For large values of $\gamma$, the norm quickly goes to zero as
gradient is propagated back in time.  For small values of $\gamma$ the norm is
nearly constant.

To demonstrate what we think is the cause of this vanishing, we drew samples
$x$ from a set of centered Gaussian distributions with standard deviation
ranging from 0 to 1, and computed the derivative $\tanh'(x) = 1 - \tanh^2(x)
\in [0, 1]$ for each.  Figure~\ref{fig:tanh_grad} shows the empirical
distribution of the derivative as a function of standard deviation.  When the
input standard deviation is low, the input tends to be close to the origin
where the derivative is close to 1.  As the standard deviation increases, the
expected derivative decreases as the input is more likely to be in the
saturation regime.  At unit standard deviation, the expected derivative is much
smaller than 1.

We conjecture that this is what causes the gradient to vanish, and recommend
initializing $\gamma$ to a small value.  In our trials we found that values of
0.01 or lower caused instabilities during training.  Our choice of 0.1 seems to
work well across different tasks.

\section{Experiments}
\label{sec:experiments}

This section presents an empirical evaluation of the proposed batch-normalized
LSTM on four different tasks.  Note that for all the experiments, we initialize
the batch normalization scale and shift parameters $\gamma$ and $\beta$ to
$0.1$ and $0$ respectively.

\subsection{Sequential MNIST}
\label{sec:seqmnist}

We evaluate our batch-normalized LSTM on a sequential version of the MNIST
classification task~\citep{le2015simple}.  The model processes each image one
pixel at a time and finally predicts the label.  We consider both sequential
MNIST tasks, MNIST and permuted MNIST ($p$MNIST).  In MNIST, the pixels are
processed in scanline order.  In $p$MNIST the pixels are processed in a fixed
random order.

Our baseline consists of an LSTM with 100 hidden units, with a softmax
classifier to produce a prediction from the final hidden state.  We use
orthogonal initialization for all weight matrices, except for the
hidden-to-hidden weight matrix which we initialize to be the identity matrix,
as this yields better generalization performance on this task for both models.
The model is trained using RMSProp~\citep{rmsprop} with learning rate of
$10^{-3}$ and $0.9$ momentum.  We apply gradient clipping at 1 to avoid
exploding gradients.

The in-order MNIST task poses a unique problem for our model: the input for the
first hundred or so timesteps is constant across examples since the upper
pixels are almost always black.  This causes the variance of the hidden states
to be exactly zero for a long period of time.  Normalizing these zero-variance
activations involves dividing zero by a small number at many timesteps, which
does not affect the forward-propagated activations but causes the
back-propagated gradient to explode.  We work around this by adding Gaussian
noise to the initial hidden states.  Although the normalization amplifies the
noise to signal level, we find that it does not hurt performance compared to
data-dependent ways of initializing the hidden states.

\begin{figure}[!t]
\center
\includegraphics[width=6.7cm]{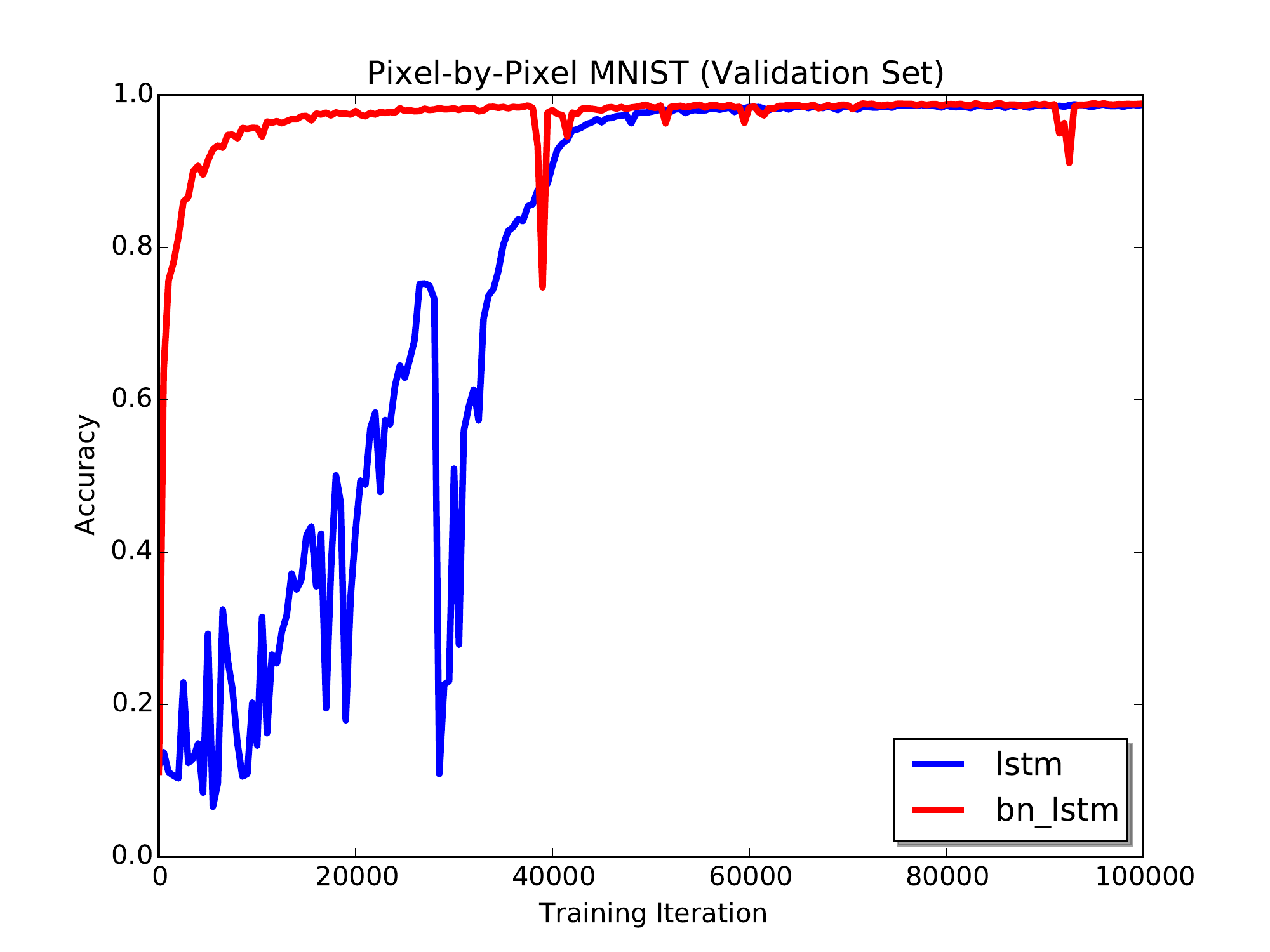}
\includegraphics[width=6.7cm]{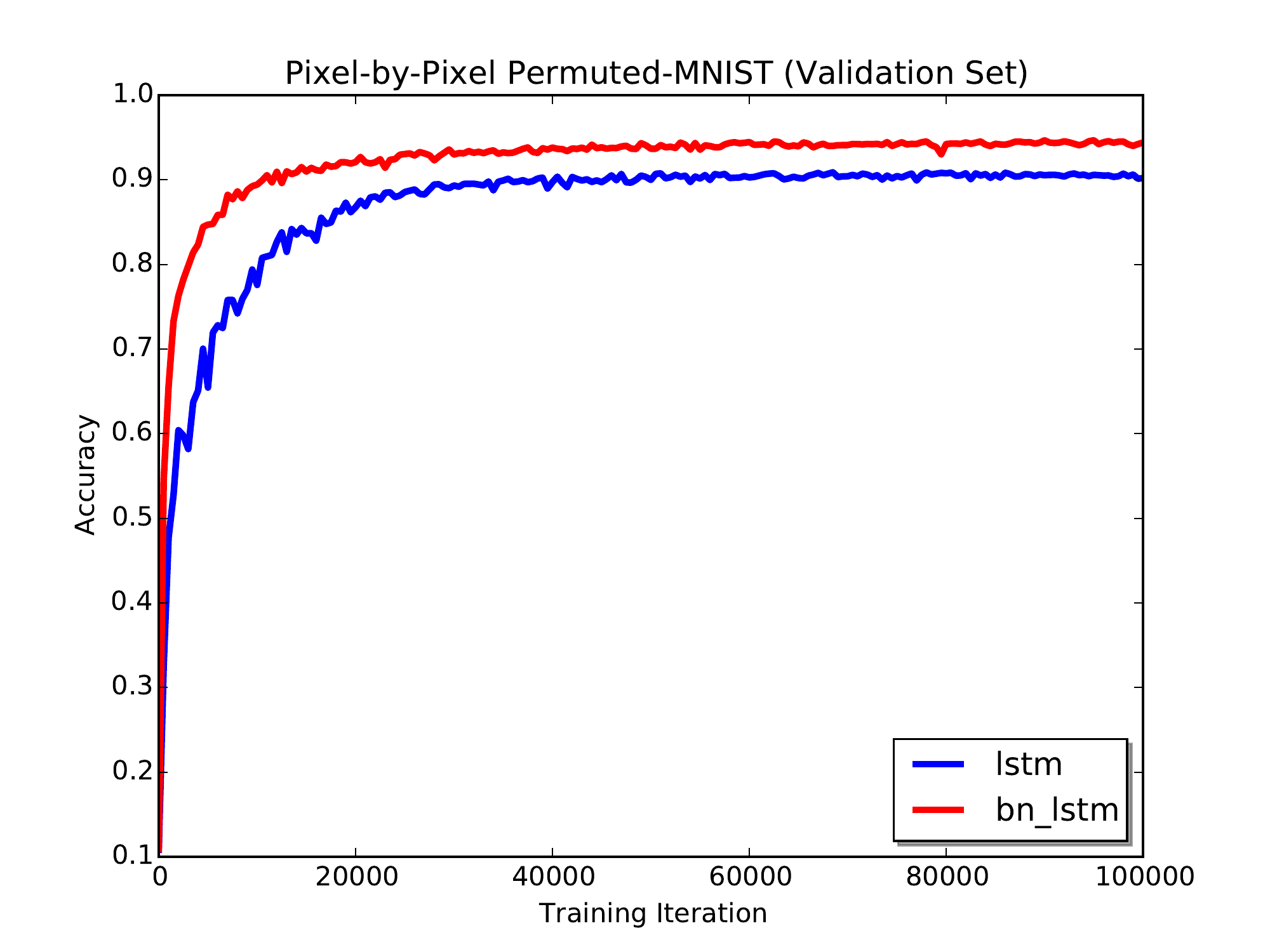}
\caption{Accuracy on the validation set for the pixel by pixel MNIST
classification tasks. The batch-normalized LSTM is able to converge faster
relatively to a baseline LSTM.  Batch-normalized  LSTM also shows some improve
generalization on the permuted sequential MNIST that require to preserve
long-term memory information.}
\label{fig:seqmnist_valid}
\end{figure}

\begin{table}[!hb]
\center
\begin{tabular}{@{}lcc@{}}
  \toprule
  \bf Model & \bf MNIST & \bf $p$MNIST \\
  \midrule
  TANH-RNN~\citep{le2015simple} & 35.0 & 35.0\\
  $i$RNN~\citep{le2015simple} & 97.0 & 82.0\\
  $u$RNN~\citep{urnn} & 95.1 & 91.4\\
  $s$TANH-RNN~\citep{zhang2016architectural} & 98.1 & 94.0\\
  \midrule
  LSTM (ours) & 98.9 & 90.2\\
  BN-LSTM (ours) & \textbf{99.0} & \textbf{95.4}\\
  \bottomrule
\end{tabular}
\caption{Accuracy obtained on the test set for the pixel by pixel MNIST classification tasks}
\label{tab:seqmnist_test}

\end{table}

In Figure~\ref{fig:seqmnist_valid} we show the validation accuracy while
training for both LSTM and batch-normalized LSTM (BN-LSTM).  BN-LSTM converges
faster than LSTM on both tasks.  Additionally, we observe that BN-LSTM
generalizes significantly better on $p$MNIST.  It has been highlighted
in~\cite{urnn} that $p$MNIST contains many longer term dependencies across
pixels than in the original pixel ordering, where a lot of structure is local.
A recurrent network therefore needs to characterize dependencies across varying
time scales in order to solve this task.  Our results suggest that BN-LSTM is
better able to capture these long-term dependencies.

Table~\ref{tab:seqmnist_test} reports the test set accuracy of the early stop
model for LSTM and BN-LSTM using the population statistics.  Recurrent batch
normalization leads to a better test score, especially for $p$MNIST where
models have to leverage long-term temporal depencies.  In addition,
Table~\ref{tab:seqmnist_test} shows that our batch-normalized LSTM achieves
state of the art on both MNIST and $p$MNIST.

\subsection{Character-level Penn Treebank}
\begin{table}
  \center
\begin{tabular}{@{}lc@{}}
  \toprule
  \bf Model & \bf Penn Treebank \\
  \midrule
  LSTM~\citep{graves2013generating} &  1.26\footnote{Our performance does not directly compare against~\citet{graves2013generating} as they use a different dataset split.}\\
  \midrule
  HF-MRNN~\citep{mikolov2012subword} & 1.41 \\
  Norm-stabilized LSTM~\citep{krueger} & 1.39 \\
  ME n-gram~\citep{mikolov2012subword} & 1.37 \\
  \midrule
  LSTM (ours) & 1.38 \\
  BN-LSTM (ours) & 1.32 \\
  \midrule
  Zoneout~\citep{krueger2016zoneout} & 1.27 \\
  HM-LSTM~\citep{chung2016hierarchical} & 1.24 \\
  HyperNetworks~\citep{ha2016hypernetworks} & \textbf{1.22} \\
  \bottomrule
\end{tabular}
\caption{Bits-per-character on the Penn Treebank test sequence.}
\label{tab:ptb_test}
\end{table}

We evaluate our model on the task of character-level language modeling on the
Penn Treebank corpus~\citep{penntreebank} according to the train/valid/test
partition of~\citet{mikolov2012subword}.  For training, we segment the training
sequence into examples of length 100.  The training sequence does not cleanly
divide by 100, so for each epoch we randomly crop a subsequence that does and
segment that instead.

Our baseline is an LSTM with 1000 units, trained to predict the next character
using a softmax classifier on the hidden state $\vect{h}_t$.  We use stochastic
gradient descent on minibatches of size 64, with gradient clipping at 1.0 and
step rule determined by Adam~\citep{kingma2014adam} with learning rate 0.002.
We use orthogonal initialization for all weight matrices.  The setup for the
batch-normalized LSTM is the same in all respects except for the introduction
of batch normalization as detailed in~\ref{sec:recurrent-batch-normalization}.

\begin{figure}[!hb]
  \center%
  \subfigure[
    Performance in bits-per-character on length-100 subsequences of the Penn Treebank
    validation sequence during training.
]{%
    \label{fig:ptb_valid}
    \includegraphics[width=.45\textwidth]{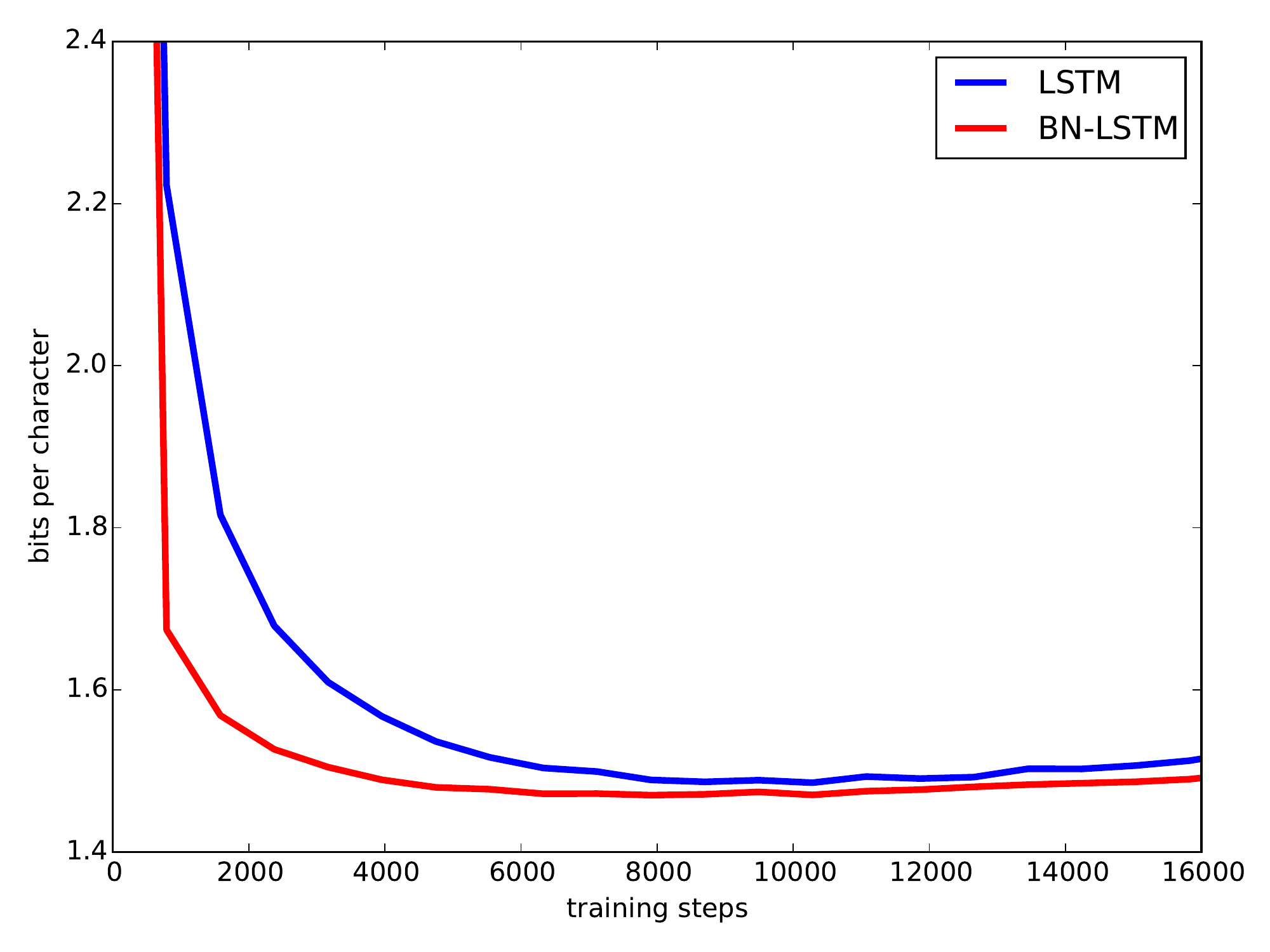}
  }%
  \hspace{2mm}%
  \subfigure[
    Generalization to longer subsequences of Penn Treebank using population statistics.
    The subsequences are taken from the test sequence.
]{%
    \label{fig:ptb_lengths}
    \includegraphics[width=.45\textwidth]{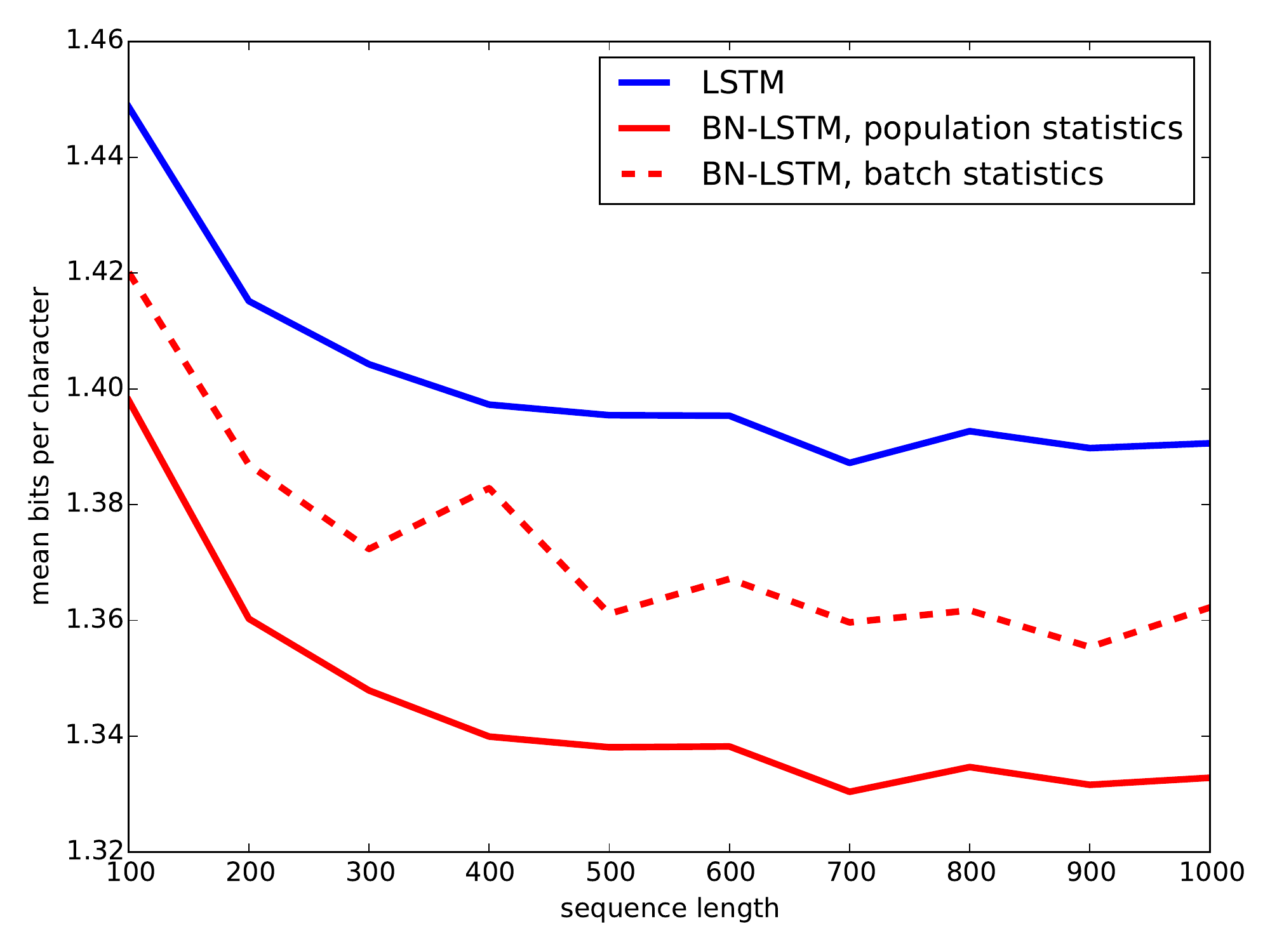}
  }
  \caption{Penn Treebank evaluation}
  \label{fig:ptb}
\end{figure}

We show the learning curves in Figure~\ref{fig:ptb_valid}.  BN-LSTM converges
faster and generalizes better than the LSTM baseline.
Figure~\ref{fig:ptb_lengths} shows the generalization of our model to longer
sequences.  We observe that using the population statistics improves
generalization performance, which confirms that repeating the last population
statistic (cf. Section~\ref{sec:recurrent-batch-normalization}) is a viable
strategy.  In table~\ref{tab:ptb_test} we report the performance of our best
models (early-stopped on validation performance) on the Penn Treebank test
sequence.  Follow up works havd since improved the state of the
art~\citep{krueger2016zoneout,chung2016hierarchical,ha2016hypernetworks}.

\subsection{Text8}

We evaluate our model on a second character-level language modeling task on the
much larger text8 dataset~\citep{mahoney2009large}.  This dataset is derived
from Wikipedia and consists of a sequence of 100M characters including only
alphabetical characters and spaces.  We
follow~\citet{mikolov2012subword,zhang2016architectural} and use the first 90M
characters for training, the next 5M for validation and the final 5M characters
for testing.  We train on nonoverlapping sequences of length 180.

Both our baseline and batch-normalized models are LSTMs with 2000 units,
trained to predict the next character using a softmax classifier on the hidden
state $\vect{h}_t$. We use stochastic gradient descent on minibatches of size
128, with gradient clipping at 1.0 and step rule determined by
Adam~\citep{kingma2014adam} with learning rate 0.001.  All weight matrices were
initialized to be orthogonal.

We early-stop on validation performance and report the test performance of the
resulting model in table~\ref{tab:text8_test}.  We observe that BN-LSTM obtains
a significant performance improvement over the LSTM baseline.
\citet{chung2016hierarchical} has since improved on our performance.

\begin{table}[!hb]
  \center
  \begin{tabular}{@{}lc@{}}
  \toprule
  \bf Model & \bf text8 \\
  \midrule
  $td$-LSTM~\citep{zhang2016architectural} & 1.63 \\
  HF-MRNN~\citep{mikolov2012subword} & 1.54 \\
  skipping RNN~\citep{pachitariu2013regularization} & 1.48 \\
  \midrule
  LSTM (ours) &  1.43 \\
  BN-LSTM (ours) & 1.36 \\
  \midrule
  HM-LSTM~\citep{chung2016hierarchical} & \textbf{1.29} \\
  \bottomrule
\end{tabular}
\caption{Bits-per-character on the text8 test sequence.}
\label{tab:text8_test}
\end{table}

\subsection{Teaching Machines to Read and Comprehend}
\label{sec:less-attr}

Recently,~\citet{attentivereader} introduced a set of challenging benchmarks
for natural language processing, along with neural network architectures to
address them.  The tasks involve reading real news articles and answering
questions about their content.  Their principal model, the Attentive Reader, is
a recurrent neural network that invokes an attention mechanism to locate
relevant information in the document.  Such models are notoriously hard to
optimize and yet increasingly popular.

To demonstrate the generality and practical applicability of our proposal,
we apply batch normalization in the Attentive Reader model
and show that this drastically improves training.

\begin{figure}[!ht]
  \center%
  \subfigure[
Error rate on the validation set for the Attentive Reader models on a variant
of the CNN QA task~\citep{attentivereader}.  As detailed in Appendix~\ref{sec:more-attr}, the
theoretical lower bound on the error rate on this task is 43\%.
]{%
    \label{fig:attr_valid2}
    \includegraphics[width=.45\textwidth]{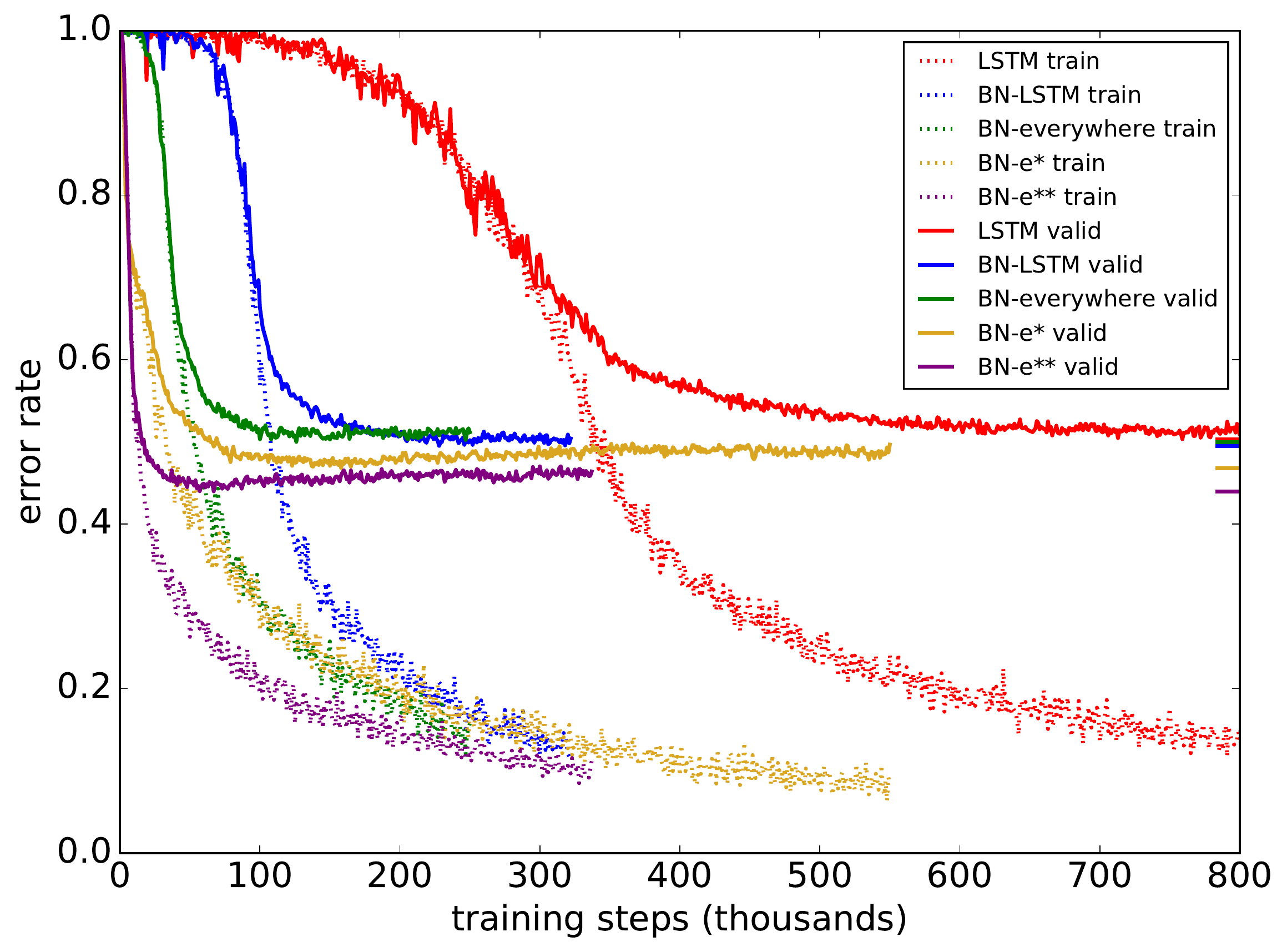}
  }%
  \hspace{2mm}%
  \subfigure[
Error rate on the validation set on the full CNN QA task from~\citet{attentivereader}.
]{%
    \label{fig:attr_full_valid}
    \includegraphics[width=.45\textwidth]{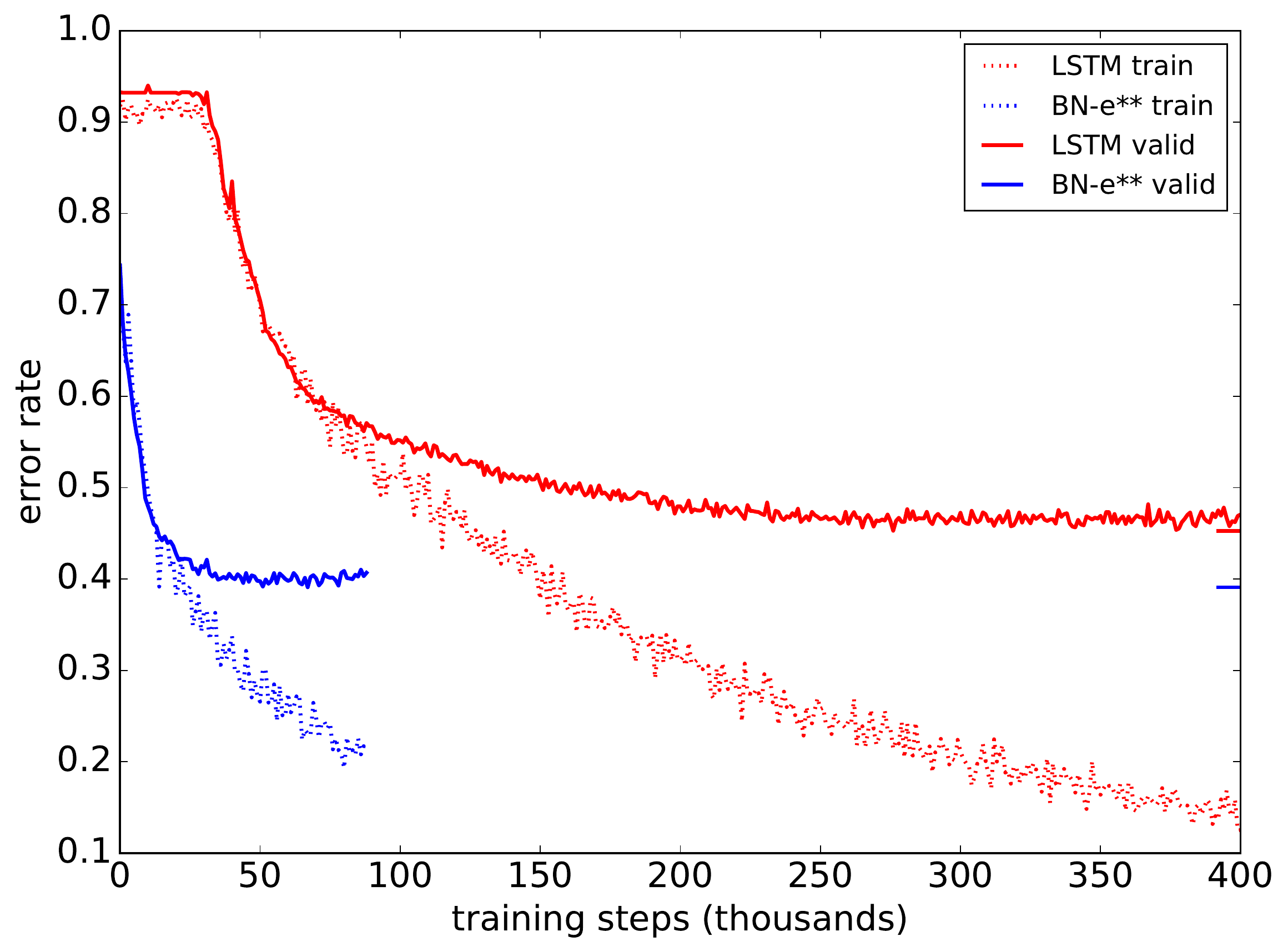}
  }
  \caption{Training curves on the CNN question-answering tasks.}
  \label{fig:attr2}
\end{figure}

We evaluate several variants.  The first variant, referred to as BN-LSTM,
consists of the vanilla Attentive Reader model with the LSTM simply replaced by
our BN-LSTM reparameterization.  The second variant, termed BN-everywhere, is
exactly like the first, except that we also introduce batch normalization into
the attention computations, normalizing each term going into the $\tanh$
nonlinearities.

Our third variant, BN-e*, is like BN-everywhere, but improved to more carefully
handle variable-length sequences.  Throughout this experiment we followed the
common practice of padding each batch of variable-length data with zeros.
However, this biases the batch mean and variance of $\vect{x}_t$ toward zero.
We address this effect using \emph{sequencewise} normalization of the inputs as
proposed by~\citet{cesar,baidu}.  That is, we share statistics over time for
normalization of the input terms $\mat{W}_x \vect{x}_t$, but \emph{not} for the
recurrent terms $\mat{W}_h \vect{h}_t$ or the cell output $\vect{c}_t$.  Doing
so avoids many issues involving degenerate statistics due to input sequence
padding.

Our fourth and final variant BN-e** is like BN-e* but bidirectional.  The main
difficulty in adapting to bidirectional models also involves padding.  Padding
poses no problem as long as it is properly ignored (by not updating the hidden
states based on padded regions of the input).  However to perform the reverse
application of a bidirectional model, it is common to simply reverse the padded
sequences, thus moving the padding to the front.  This causes similar problems
as were observed on the sequential MNIST task (Section~\ref{sec:seqmnist}): the
hidden states will not diverge during the initial timesteps and hence their
variance will be severely underestimated.  To get around this, we reverse only
the unpadded portion of the input sequences and leave the padding in place.

See Appendix~\ref{sec:more-attr} for hyperparameters and task details.

Figure~\ref{fig:attr_valid2} shows the learning curves for the different
variants of the attentive reader.  BN-LSTM trains dramatically faster than the
LSTM baseline.  BN-everywhere in turn shows a significant improvement over
BN-LSTM.  In addition, both BN-LSTM and BN-everywhere show a generalization
benefit over the baseline.  The validation curves have minima of 50.3\%, 49.5\%
and 50.0\% for the baseline, BN-LSTM and BN-everywhere respectively.  We
emphasize that these results were obtained without any tweaking -- all we did
was to introduce batch normalization.

BN-e* and BN-e** converge faster yet, and reach lower minima: 47.1\% and 43.9\%
respectively.

\begin{table}[!hb]
  \center
  \begin{tabular}{@{}lrr@{}}
  \toprule
  \bf Model & \bf CNN valid & \bf CNN test \\
  \midrule
  Attentive Reader~\citep{attentivereader} & 38.4 & 37.0 \\
  \midrule
  LSTM (ours) & 45.5 & 45.0 \\
  BN-e** (ours) & \textbf{37.9} & \textbf{36.3} \\
  \bottomrule
\end{tabular}
\caption{Error rates on the CNN question-answering task~\citet{attentivereader}.}
\label{tab:attr_full}
\end{table}

We train and evaluate our best model, BN-e**, on the full task from~\citep{attentivereader}.
On this dataset we had to reduce the number of hidden units to 120 to avoid severe overfitting.
Training curves for BN-e** and a vanilla LSTM are shown in Figure~\ref{fig:attr_full_valid}.
Table~\ref{tab:attr_full} reports performances of the early-stopped models.

\section{Conclusion}

Contrary to previous findings by~\citet{cesar,baidu}, we have demonstrated that
batch-normalizing the hidden states of recurrent neural networks greatly
improves optimization.  Indeed, doing so yields benefits similar to those of
batch normalization in feed-forward neural networks: our proposed BN-LSTM
trains faster and generalizes better on a variety of tasks including language
modeling and question-answering.  We have argued that proper initialization of
the batch normalization parameters is crucial, and suggest that previous
difficulties~\citep{cesar, baidu} were due in large part to improper
initialization.  Finally, we have shown our model to apply to complex settings
involving variable-length data, bidirectionality and highly nonlinear attention
mechanisms.

\section*{Acknowledgements}

The authors would like to acknowledge the following agencies for
research funding and computing support: the Nuance Foundation, Samsung, NSERC, Calcul Qu\'{e}bec, Compute Canada,
the Canada Research Chairs and CIFAR.
Experiments were carried out using the Theano~\citep{theano} and the
Blocks and Fuel~\citep{blocks} libraries for scientific computing.
We thank David Krueger, Saizheng Zhang, Ishmael Belghazi and Yoshua Bengio for discussions and suggestions.

\bibliography{index}
\bibliographystyle{iclr2017_conference}

\newpage

\appendix

\section{Convergence of population statistics} \label{sec:popstat_stationarity}

\begin{figure}[!h]
\center
\includegraphics[width=0.8\textwidth]{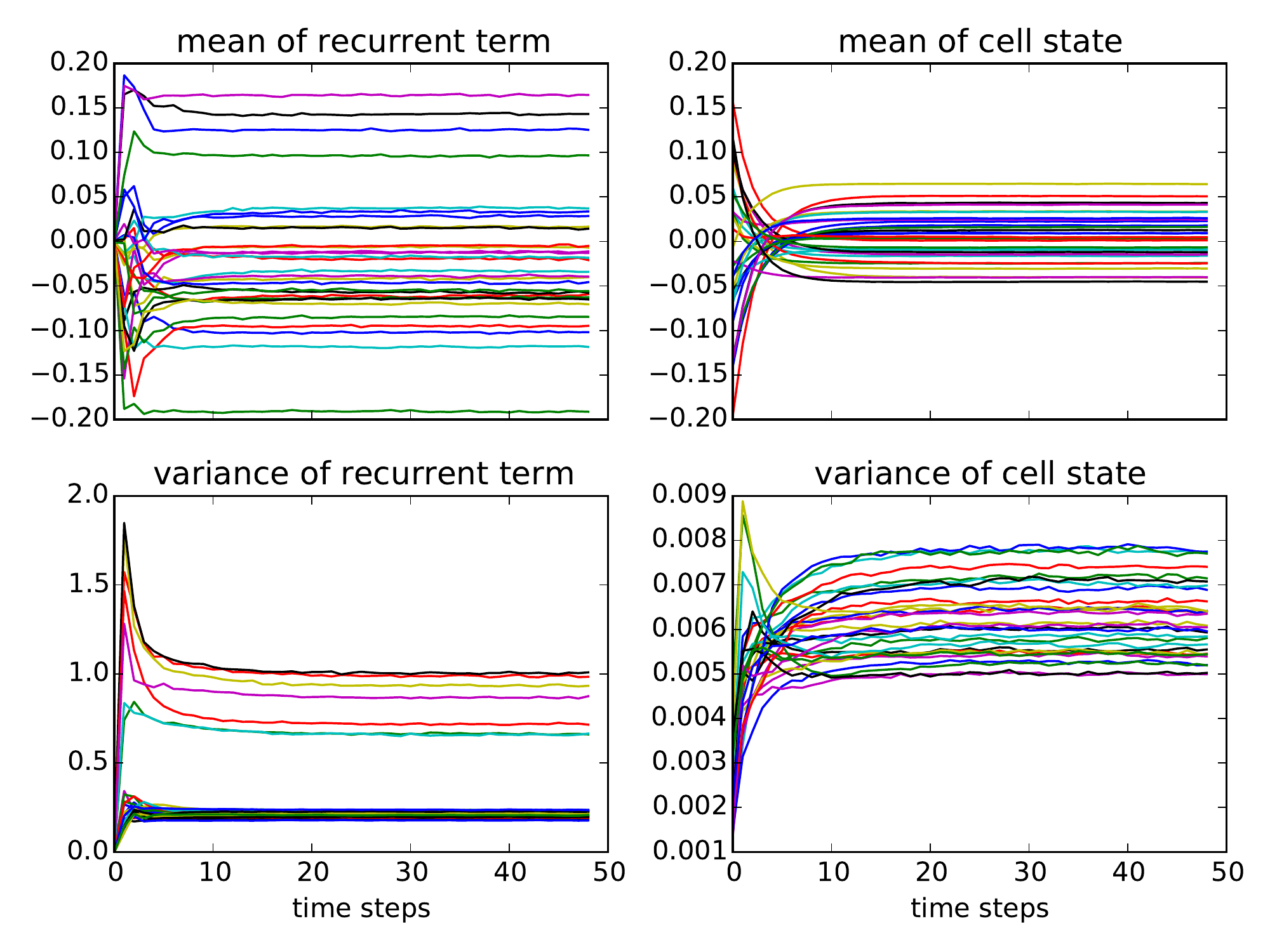}
\caption{Convergence of population statistics to stationary distributions on the 
Penn Treebank task. The horizontal axis denotes RNN time. Each curve corresponds to a single hidden unit. Only a random subset of units is shown.
See Section~\ref{sec:recurrent-batch-normalization} for discussion.}
\label{fig:popstat_stationarity}
\end{figure}

\section{Sensitivity to initialization of $\gamma$}

\begin{figure}[!h]
\center
\includegraphics[width=0.9\textwidth]{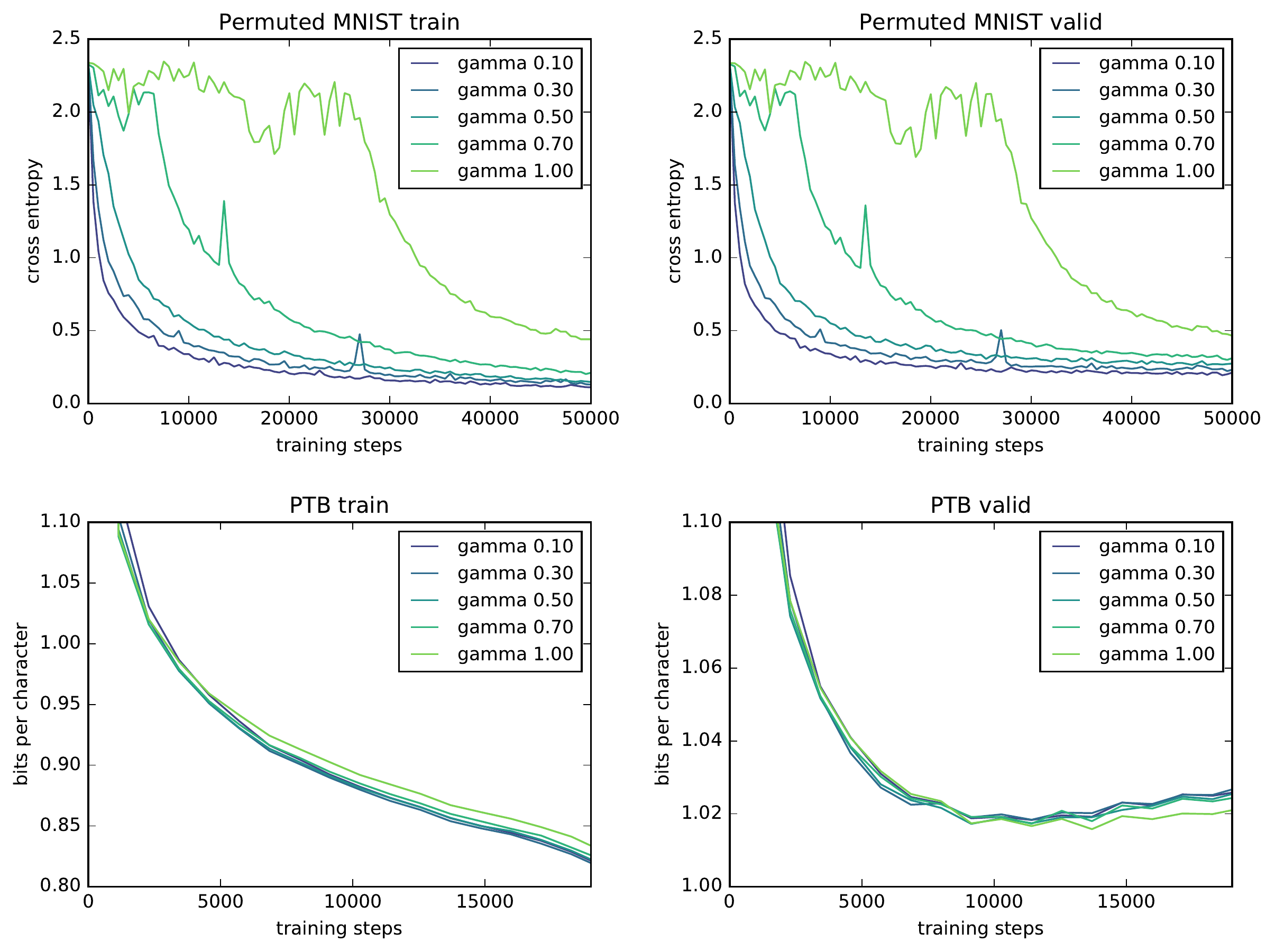}
\caption{Training curves on $p$MNIST and Penn Treebank for various initializations of $\gamma$.}
\label{fig:gammas}
\end{figure}

In Section~\ref{sec:activation-variance} we investigated the effect of initial $\gamma$ on gradient flow.
To show the practical implications of this, we performed several experiments on the $p$MNIST and Penn Treebank benchmarks.
The resulting performances are shown in Figure~\ref{fig:gammas}.

The $p$MNIST training curves confirm that higher initial values of $\gamma$ are detrimental to the optimization of the model.
For the Penn Treebank task however, the effect is gone.

We believe this is explained by the difference in the nature of the two tasks.
For $p$MNIST, the model absorbs the input sequence and only at the end of the sequence does it make a prediction on which it receives feedback.
Learning from this feedback requires propagating the gradient all the way back through the sequence.

In the Penn Treebank task on the other hand, the model makes a prediction at each timestep.
At each step of the backward pass, a fresh learning signal is added to the backpropagated gradient.
Essentially, the model is able to get off the ground by picking up short-term dependencies.
This fails on $p$MNIST wich is dominated by long-term dependencies~\citep{urnn}.

\section{Teaching Machines to Read and Comprehend: Task setup} \label{sec:more-attr}

We evaluate the models on the question answering task using the CNN corpus~\citep{attentivereader},
with placeholders for the named entities. We follow a similar preprocessing
pipeline as~\citet{attentivereader}.  During training, we randomly sample the
examples with replacement and shuffle the order of the placeholders in each
text inside the minibatch.  We use a vocabulary of 65829 words.

We deviate from~\citet{attentivereader} in order to save computation:
we use only the 4 most relevant sentences from the description,
as identified by a string matching procedure.
Both the training and validation sets are preprocessed in this way.
Due to imprecision this heuristic sometimes strips the answers from the passage,
putting an upper bound of 57\% on the validation accuracy that can be achieved.

For the reported performances, the first three models (LSTM, BN-LSTM and BN-everywhere) are trained using the
exact same hyperparameters, which were chosen because they work well for the
baseline.  The hidden state is composed of 240 units.  We use stochastic
gradient descent on minibatches of size 64, with gradient clipping at 10 and
step rule determined by Adam~\citep{kingma2014adam} with learning rate $8
\times 10^{-5}$.

For BN-e* and BN-e**, we use the same hyperparameters except that we reduce the
learning rate to $8 \times 10^{-4}$ and the minibatch size to 40.

\section{Hyperparameter Searches}

Table~\ref{tab:hyperparams} reports hyperparameter values that were tried in the experiments.

\begin{table}[!hb]

  \subtable[MNIST and $p$MNIST]
  {
    \begin{tabular}{ll}
      Learning rate: & 1e-2, 1e-3, 1e-4\\
      RMSProp momentum: & 0.5,  0.9\\
      Hidden state size: & 100, 200, 400\\
      Initial $\gamma$: & 1e-1, 3e-1, 5e-1, 7e-1, 1.0\\
    \end{tabular}
  }
    \subtable[Penn Treebank]
  {
    \begin{tabular}{ll}
      Learning rate: & 1e-1, 1e-2, 2e-2, 1e-3\\
      Hidden state size: & 800, 1000, 1200, 1500, 2000\\
      Batch size: & 32, 64, 100, 128\\
      Initial $\gamma$: & 1e-1, 3e-1, 5e-1, 7e-1, 1.0\\
    \end{tabular}
  }
    \subtable[Text8]
  {
    \begin{tabular}{ll}
      Learning rate: & 1e-1, 1e-2, 1e-3\\
      Hidden state size: & 500, 1000, 2000, 4000\\
    \end{tabular}
  }
    \subtable[Attentive Reader]
  {
    \begin{tabular}{ll}
      Learning rate: & 8e-3, 8e-4, 8e-5, 8e-6\\
      Hidden state size: & 60, 120, 240, 280\\
    \end{tabular}
  }

  \caption{Hyperparameter values that have been explored in the experiments.}
  \label{tab:hyperparams}

\end{table}

For MNIST and $p$MNIST, the hyperparameters were varied independently.
For Penn Treebank, we performed a full grid search on learning rate and hidden state size, and later performed a sensitivity analysis on the batch size and initial $\gamma$.
For the text8 task and the experiments with the Attentive Reader, we carried out a grid search on the learning rate and hidden state size.

The same values were tried for both the baseline and our BN-LSTM.
In each case, our reported results are those of the model with the best validation performance.

\end{document}